\documentclass[runningheads]{llncs}

\usepackage{eccv}

\usepackage{eccvabbrv}

\usepackage{graphicx}
\usepackage{booktabs}

\usepackage[accsupp]{axessibility}  

\usepackage[pagebackref,breaklinks,colorlinks,citecolor=eccvblue]{hyperref}

\usepackage{orcidlink}

\usepackage{enumitem} 
\usepackage{wrapfig}

\begin{document}

\title{ReImagine: Rethinking Controllable High-Quality \\ Human Video Generation via \\ Image-First Synthesis} 

\titlerunning{ReImagine: Image-First Human Video Generation}

\author{
Zhengwentai Sun\inst{1,2} \and
Keru Zheng\inst{1} \and
Chenghong Li\inst{1,2} \and
Hongjie Liao\inst{1} \and \\
Xihe Yang\inst{1} \and
Heyuan Li\inst{1} \and
Yihao Zhi\inst{1,2} \and
Shuliang Ning\inst{1} \and \\
Shuguang Cui\inst{1,2} \and
Xiaoguang Han\inst{1,2}\thanks{Corresponding author: \email{hanxiaoguang@cuhk.edu.cn}}
}

\authorrunning{Z. Sun et al.}

\institute{
School of Science and Engineering, The Chinese University of Hong Kong, Shenzhen \and
Future Network of Intelligence Institute, CUHK-Shenzhen, China
}

\maketitle

\begin{figure}[t]
    \centering
    \includegraphics[width=.89\linewidth]{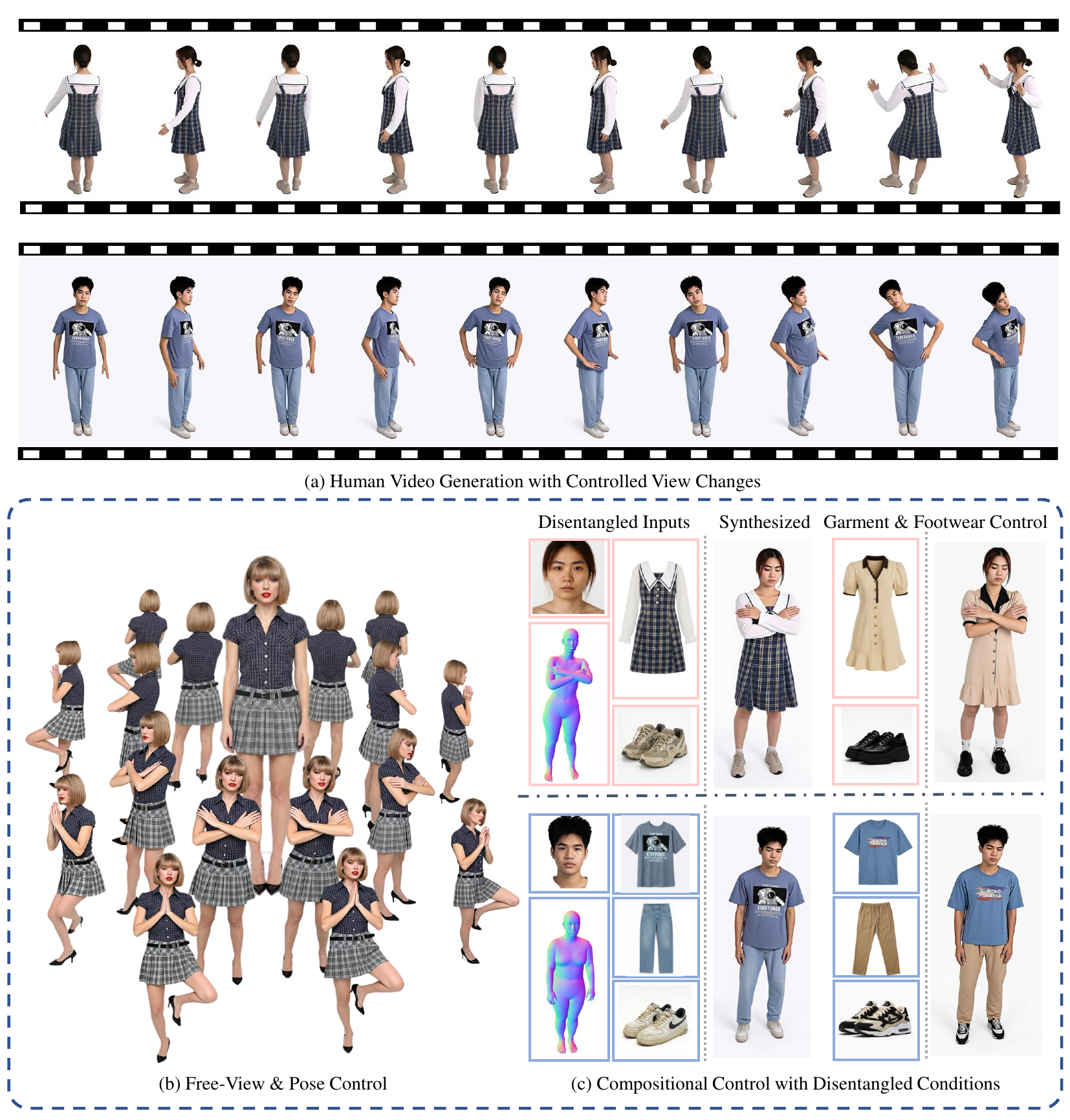}
    \caption{
Our method enables controllable human synthesis at multiple levels.
(a) Our pipeline generates temporally coherent videos with explicit control over body pose and camera viewpoint.
(b) Our image model generalizes to in-the-wild references, producing diverse poses and viewpoints with consistent appearance.
(c) As an additional contribution, our end-to-end model supports compositional synthesis with disentangled control over identity, garments, footwear, and pose.
}
    \label{fig:teaser}
    \vspace{-18px}
\end{figure}

\begin{abstract}
  Human video generation remains challenging due to the difficulty of jointly modeling human appearance, motion, and camera viewpoint under limited multi-view data. Existing methods often address these factors separately, resulting in limited controllability or reduced visual quality. We revisit this problem from an image-first perspective, where high-quality human appearance is learned via image generation and used as a prior for video synthesis, decoupling appearance modeling from temporal consistency. We propose a pose- and viewpoint-controllable pipeline that combines a pretrained image backbone with SMPL-X-based motion guidance, together with a training-free temporal refinement stage based on a pretrained video diffusion model. Our method produces high-quality, temporally consistent videos under diverse poses and viewpoints. We also release a canonical human dataset and an auxiliary model for compositional human image synthesis. Code and data are publicly available at \url{https://github.com/Taited/ReImagine}.
\end{abstract}

\section{Introduction}
\label{sec:intro}

Human video generation is an active research topic in computer vision and computer graphics, with applications in virtual reality, virtual try-on, and digital content creation. Recent advances in generative models~\cite{wan2025wan, wang2025unianimate, cheng2025wananimateunifiedcharacteranimation} have significantly improved the realism of synthesized human appearance and motion. However, achieving simultaneous control over body pose, human appearance, and camera viewpoint remains challenging. This is due to both modeling complexity and data limitations. Learning these factors jointly typically requires large-scale, high-quality multi-view video data, which is difficult to obtain in practice. As a result, existing approaches often address these aspects separately, and their joint modeling remains limited.

Existing methods reflect this fragmentation. Pose-controllable approaches~\cite{hu2024animate, xu2024magicanimate} rely on 2D skeleton sequences and produce strong results under fixed viewpoints, while Champ~\cite{zhu2024champ} improves motion consistency by introducing 3D pose guidance through parametric SMPL representations. However, these methods do not explicitly model viewpoint variation. Scaling to large video generation models does not fully resolve this limitation. 
Even models equipped with pose conditioning~\cite{cheng2025wan, Wan2.1FunV1.1_14B_Control} provide limited control over camera viewpoint. In contrast, human-centric frameworks such as HuMo~\cite{chen2025humo} and Phantom~\cite{liu2025phantom} rely on audio or multi-subject inputs, rather than structured geometric control over pose and viewpoint.
From the viewpoint perspective, MV-Performer~\cite{zhi2025mv} enables novel-view synthesis from monocular input but does not support explicit pose control. 
Human4DiT~\cite{shao2024human4dit} is closest to our goal, performing pose-driven multi-view 4D generation. However, it is trained on web-scraped monocular videos, where viewpoint variation is not explicitly controlled during training. As a result, achieving precise pose- and viewpoint-controllable generation remains challenging.

A fundamental limitation behind these approaches is the availability of data. Training a video generation model with joint pose and viewpoint control, for example in the form of paired SMPL-X motion and multi-view video data, is straightforward. However, such data remains scarce in both scale and quality. Even the largest existing multi-view human video datasets, such as MVHumanNet++~\cite{li2025mvhumannet++}, remain insufficient to fully support high-quality video generation. Under these conditions, directly training video models often leads to a trade-off between controllability and visual fidelity.

\begin{figure}[t]
    \centering
    \includegraphics[width=0.99\linewidth]{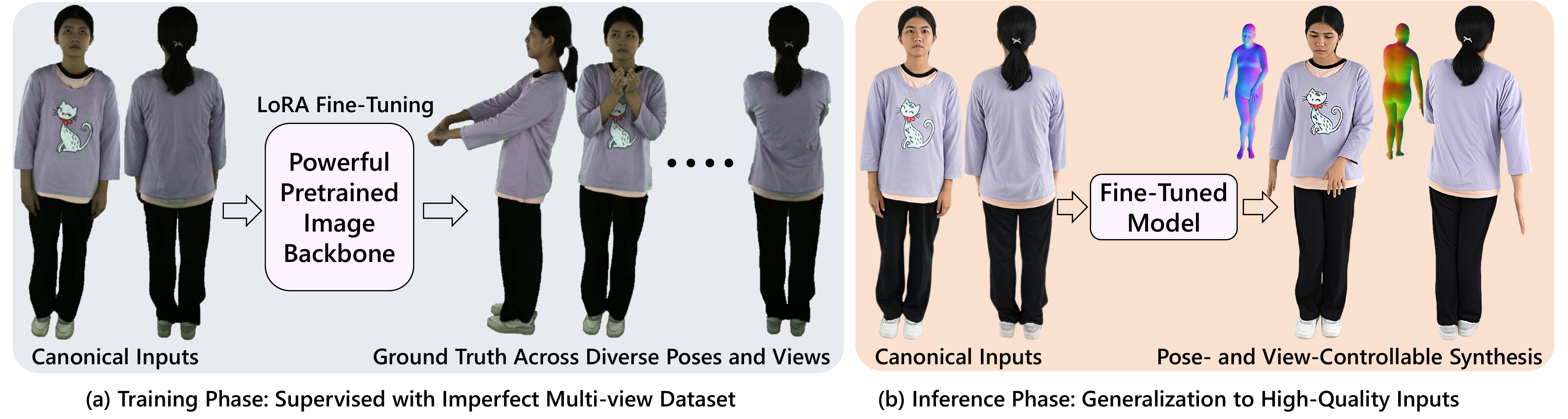}
    \caption{
Overview of our image-first training and inference paradigm. 
(a) During training, a powerful pretrained image backbone is fine-tuned via lightweight LoRA adaptation using an imperfect multi-view dataset with structured pose and viewpoint supervision. 
(b) At inference, the fine-tuned model generalizes to high-quality canonical inputs and enables pose- and viewpoint-controllable human synthesis. 
This highlights our key insight that strong pretrained image priors substantially reduce the need for large-scale, high-fidelity multi-view video data.
}
    \label{fig:inputs_quality}
\end{figure}

To address this issue, we revisit controllable human video generation from an image-first perspective. 
We term our method \textbf{ReImagine}.
Specifically, we prioritize learning high-quality human appearance through image generation and use it as a strong prior for video synthesis.
ReImagine thus proceeds in two stages: a pose- and viewpoint-guided image synthesis stage, followed by a training-free temporal consistency stage.
Modern pretrained image generation models \cite{labs2025flux,wu2025qwen,esser2024scaling} already encode strong priors over human appearance. 
As a result, fine-tuning such models for pose- and viewpoint-conditioned generation remains effective even with imperfect multi-view data, since the model primarily learns geometric conditioning (see Fig.~\ref{fig:inputs_quality}).
Temporal consistency can then be handled in a separate stage using a pretrained video diffusion model, without requiring task-specific video training. 
This decoupled design avoids the need to jointly learn appearance and temporal dynamics from limited data.

Concretely, given canonical front--back human images and a sequence of SMPL-X parameters, we fine-tune FLUX Kontext~\cite{labs2025flux} to generate each frame under pose and viewpoint guidance derived from rendered SMPL-X normal maps. Frame-wise generation inevitably introduces subtle appearance variations across frames, such as changes in clothing wrinkles. 
To mitigate this issue, we introduce a training-free temporal refinement stage based on a pretrained video diffusion model (Wan~\cite{wan2025wan}), combined with spatiotemporal regularization.
This stage improves temporal coherence without requiring additional video-specific training. The overall pipeline is simple, data-efficient, and compatible with existing image generation frameworks.

Our pipeline takes canonical front--back images as input, which may not always be readily available in practice. To address this, we construct a paired canonical dataset consisting of disentangled face, clothing, and footwear images synthesized using GPT-4o~\cite{hurst2024gpt}. Based on this dataset, we train an end-to-end model that generates full-body human images from disentangled components and rendered SMPL-X inputs. This dataset and model are provided as additional resources to support future research on controllable human generation.

Our code, models, and dataset are publicly available at \url{https://github.com/Taited/ReImagine}.

In summary, our contributions are as follows:
\begin{itemize}[itemsep=2pt,topsep=2pt,parsep=0pt,leftmargin=2em]
\item We propose a pose- and viewpoint-controllable human video generation pipeline from canonical front--back images and SMPL-X motion sequences, achieving strong performance in temporally coherent and viewpoint-aware synthesis.
\item We demonstrate that decoupling appearance modeling from temporal consistency, by combining a pretrained image backbone with training-free refinement using a pretrained video diffusion model, enables high-quality generation without task-specific video training under limited multi-view data.
\item We construct a canonical human dataset with disentangled face, clothing, and footwear attributes, together with an end-to-end model for attribute-consistent full-body human image generation, both publicly released to support future research.
\end{itemize}

\section{Related Works}
\subsection{Image-based Human Generation}

The evolution of image-based human generation is characterized by a significant transition from adversarial learning \cite{goodfellow2020generative,isola2017image,karras2019style,karras2020analyzing,karras2017progressive} to latent diffusion paradigms \cite{podell2023sdxl,song2020denoising,peebles2023scalable}. Early research was dominated by Generative Adversarial Networks. Pioneering StyleGAN-Human \cite{fu2022stylegan} focus on unconditional human image generation, achieving photorealism through a minimax game between a generator and a discriminator. HumanGAN \cite{Sarkar2021HumanGAN} is capable of performing all tasks related to global appearance sampling, pose transfer, part and garment transfer, as well as part sampling. Nevertheless, GANs often faced challenges with training instability and mode collapse, leading to the rise of Denoising Diffusion Probabilistic Models \cite{song2020denoising, Rombach_2022_CVPR, lipman2022flow}.

Diffusion models utilize an iterative denoising process to recover clean images from Gaussian noise, offering superior stability and diversity. And beyond unconditional synthesis, the focus has shifted toward controllable human generation, where specific attributes like pose and identity must be preserved. ControlNet \cite{zhang2023adding} emerged as a pivotal framework, enabling the injection of external conditions such as skeletal poses and edges into pre-trained models without altering their original weights. Subsequent specialized architectures like HumanSD \cite{ju2023humansd} have utilized native skeleton-guided diffusion to achieve higher semantic fidelity. HyperHuman \cite{liu2023hyperhuman} proposes a Structure-Guided Refiner to compose predicted conditions for more detailed generation at higher resolution. CFLD \cite{lu2024coarse} decouple fine-grained appearance and pose information controls at different stages, circumventing the potential overfitting problem.

A critical application of these controlled generative techniques is high-fidelity Virtual Try-On (VTON). The field has evolved from explicit warping networks toward implicit, diffusion-based mechanisms that better handle complex deformations. OOTDiffusion \cite{xu2024ootdiffusion} aligns garment features with bodies using Outfitting Fusion, improving realism and controllability without redundant warping. IDM-VTON \cite{choi2025IDM} employs a visual encoder and parallel UNet for better garment encoding, while DressCode \cite{Morelli_2022_CVPR} introduces a multi-category dataset and a Pixel-level Semantic-Aware Discriminator (PSAD) to enhance image quality. IMAGDressing-v1 \cite{shen2024imagdressing} introduces the virtual dressing (VD) task, allowing merchants to showcase garments comprehensively with flexible control over faces, poses, and scenes. Most recently, we also observe that universal large models such as FLUX Kontext \cite{labs2025flux}, Qwen Image Edit \cite{wu2025qwen}, and commercial applications like GPT-4o \cite{hurst2024gpt} and Nano Banana Pro \cite{team2023gemini} can handle these tasks seamlessly without specific training.

\subsection{Video-based Human Generation}
Beyond human-centric image synthesis, human-centric video generation \cite{chen2025humo,hu2024animate,ning20251,wang2025unianimate, men2025mimo} and editing \cite{zhong2024deco,liu2024heromaker} are gaining increasing attention. As pioneered by AnimateAnyone \cite{hu2024animate}, many works \cite{men2025mimo,xu2024magicanimate,tu2025stableanimator,yang2024showmaker,zhai2024idol,zhang2024mimicmotion} extend Latent Diffusion \cite{Rombach_2022_CVPR} and ControlNet \cite{zhang2023adding} by inserting additional temporal attention blocks. MagicAnimate \cite{xu2024magicanimate} extends diffusion to person animation, achieving smooth pose transitions and dynamic visual effects. Champ \cite{zhu2024champ} incorporates a 3D human model into latent diffusion for enhanced 3D conditioning. Human4DiT \cite{shao2024human4dit} leverages diffusion transformers to synthesize 360° coherent human videos from a single image, enabling fine-grained pose and view control. Recently, the visual quality has been further enhanced with the rise of video diffusion models \cite{wan2025wan,hong2022cogvideo,yang2024cogvideox,kong2024hunyuanvideo}. Wang and Chen \cite{wang2024replace,chen2025humo} leverage the powerful capabilities of Wan2.1 \cite{wan2025wan} for consistent human image animation. MV-Performer \cite{zhi2025mv} proposes to perform human-centric 4D novel view synthesis in a generative manner. Phantom \cite{liu2025phantom} presents a unified video generation framework for both single- and multi-subject references with a new injection strategy. These works all highlight the importance of high quality video dataset, while we we provide a new insight that we can achieve this even with low-quality video data, and without video-specific temporal training.

\begin{figure}[t]
    \centering
    \includegraphics[width=0.88\linewidth]{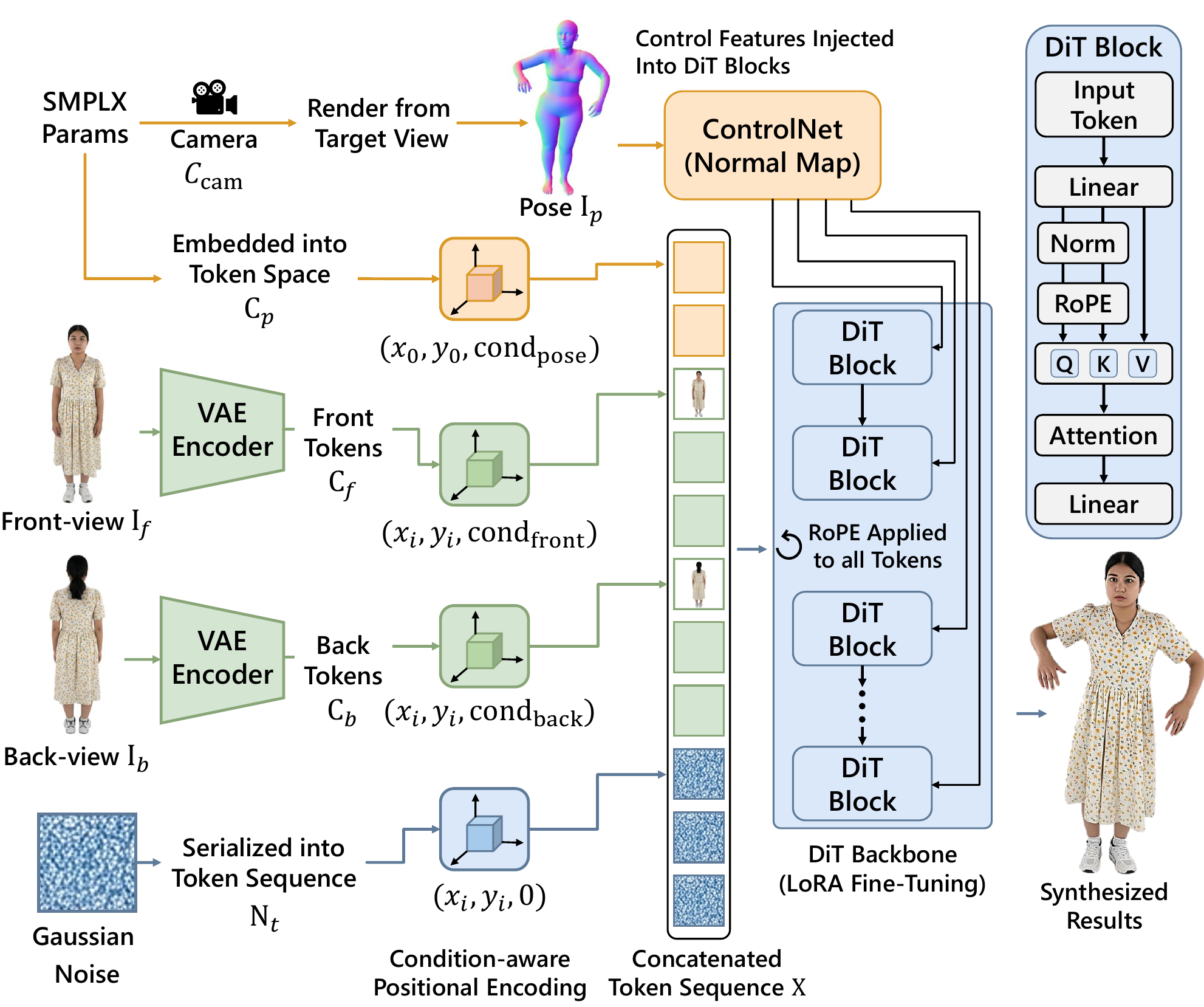}
    \caption{Overview of the proposed pose- and view-guided generation module. SMPL-X–based pose and canonical front/back appearance cues are unified in a token sequence and processed by a DiT backbone with condition-aware RoPE. The final image is obtained by decoding the generated latent using a VAE decoder (not shown for clarity).}
    \label{fig:pose_transfer}
    \vspace{-10px}
\end{figure}

\section{Method}
\vspace{-5px}
\subsection{Overview}

We propose \textbf{ReImagine}, an image-first framework for human video synthesis that decomposes the task into image generation and temporal refinement. Given canonical front-view and back-view full-body human images as appearance inputs, together with a sequence of target human poses and camera views, our method synthesizes pose- and view-consistent image frames and refines them into temporally coherent videos.

Specifically, the Pose- and View-Guided Image Synthesis Module (\cref{sec:image_module}) generates high-quality human images conditioned on structured 3D pose guidance and explicit view control. To obtain smooth video results, we further apply a Training-Free Temporal Consistency Module (\cref{sec:temporal_module}) that operates purely at inference time to suppress frame-to-frame artifacts and stabilize motion without additional training.

\subsection{Pose- and View-Guided Image Synthesis Module}
\label{sec:image_module}

We propose a pose- and view-guided generation module that incorporates 3D human geometry and canonical front–back appearance cues into a diffusion transformer (DiT). The module takes SMPL-X parameters, canonical front-view and back-view whole-body images, and Gaussian noise as input, and unifies these heterogeneous conditions in a shared token space for controllable human image synthesis (\cref{fig:pose_transfer}).

\paragraph{Condition Preparation.}
Given SMPL-X pose and shape parameters $\boldsymbol{\theta}$ and $\boldsymbol{\beta}$, we render the human mesh under target camera $\mathcal{C}_{\text{cam}}$ to obtain a surface normal map:
\begin{equation}
\mathbf{I}_p = \mathcal{R}_{\text{normal}}(\boldsymbol{\theta}, \boldsymbol{\beta}; \mathcal{C}_{\text{cam}}),
\end{equation}
where $\mathcal{R}_{\text{normal}}$ outputs per-pixel surface normals in the world coordinate system. Compared to silhouettes or depth maps, world-coordinate normals encode articulated pose and view-dependent orientation, making them suitable for joint pose–view control. The normal map is encoded by a ControlNet-style encoder and injected into DiT blocks as multi-scale pose features.

We further embed SMPL-X parameters directly into the token space via
\begin{equation}
\mathbf{C}_p = \mathrm{MLP}_{\text{pose}}([\boldsymbol{\theta}, \boldsymbol{\beta}]) \in \mathbb{R}^{N_p \times d},
\end{equation}
providing a global geometric prior.

For appearance modeling, canonical front-view and back-view images $\mathbf{I}_f$ and $\mathbf{I}_b$ are encoded by a shared VAE into tokens $\mathbf{C}_f \in \mathbb{R}^{N_f \times d}$ and $\mathbf{C}_b \in \mathbb{R}^{N_b \times d}$. This dual-view design reduces view ambiguity and supports large viewpoint changes.

At diffusion timestep $t$, stochastic noise is serialized into spatial noise tokens
\begin{equation}
\mathbf{N}_t \in \mathbb{R}^{N_z \times d}.
\end{equation}
All tokens are concatenated as
\begin{equation}
\mathbf{X} = [\mathbf{C}_p \;\Vert\; \mathbf{C}_f \;\Vert\; \mathbf{C}_b \;\Vert\; \mathbf{N}_t],
\end{equation}
and processed by the DiT backbone to predict the conditional denoising direction.

\paragraph{Condition-Aware Positional Encoding.}
To distinguish heterogeneous token types while preserving spatial structure, we introduce condition-aware positional encoding. Each token $i$ is assigned
\begin{equation}
\mathbf{p}_i = (x_i, y_i, c_i),
\end{equation}
where $(x_i, y_i)$ denote spatial coordinates and $c_i$ is a discrete condition index (pose, front-view, back-view, or noise). Image and noise tokens inherit their spatial grid coordinates, while pose tokens are assigned fixed pseudo-spatial coordinates to reflect their global, non-spatial nature.

\begin{figure}[ht]
    \centering
    \includegraphics[width=0.9\linewidth]{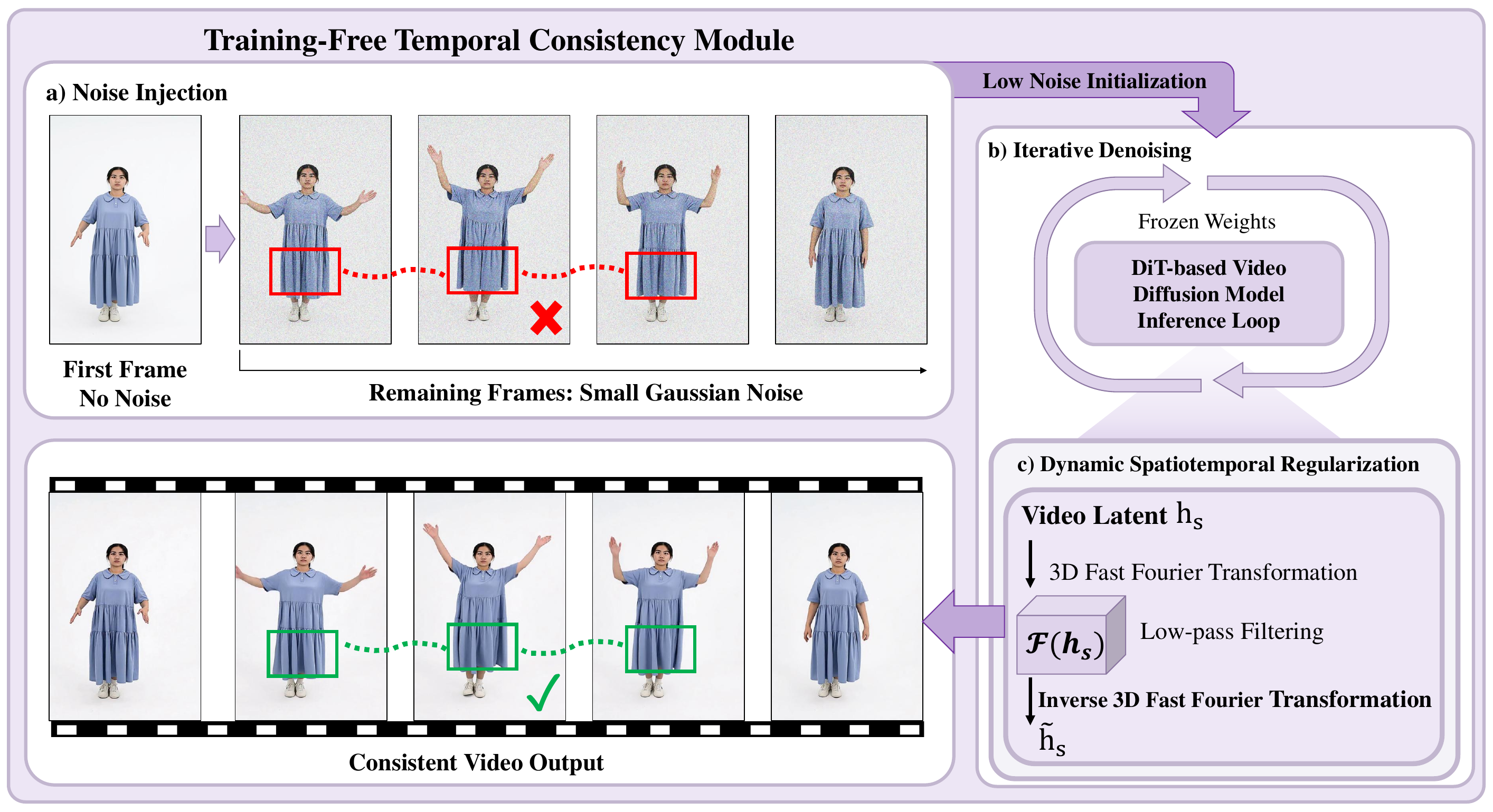}
    \caption{Training-free temporal consistency via low-noise re-denoising and spatiotemporal spectral regularization. Boxes highlight regions with improved temporal consistency.}
    \label{fig:method_temporal}
\end{figure}
We apply rotary positional embedding (RoPE) using $\mathbf{p}_i$:
\begin{equation}
\mathrm{RoPE}(\mathbf{q}_i; \mathbf{p}_i) = \mathbf{R}(\mathbf{p}_i)\,\mathbf{q}_i,
\end{equation}
with the same transformation applied to $\mathbf{k}_j$. Attention is computed as
\begin{equation}
\mathrm{Attn}(\mathbf{q}_i, \mathbf{k}_j) =
\frac{
\mathrm{RoPE}(\mathbf{q}_i; \mathbf{p}_i)^\top
\mathrm{RoPE}(\mathbf{k}_j; \mathbf{p}_j)
}{
\sqrt{d}
}.
\end{equation}
Incorporating $c_i$ enables explicit separation of pose, appearance, and noise tokens while maintaining spatial structure. RoPE is applied uniformly in every DiT block.

\paragraph{Training with Flow Matching.}
We train the model under the flow matching formulation, where the DiT predicts a conditional velocity field guided by pose and appearance conditions.

Given a clean latent $\mathbf{x}_0$ and noise $\boldsymbol{\epsilon} \sim \mathcal{N}(0, \mathbf{I})$, we define
\begin{equation}
\mathbf{x}_t = (1 - t)\,\mathbf{x}_0 + t\,\boldsymbol{\epsilon},
\quad t \sim \mathcal{U}(0,1),
\end{equation}
with target velocity
\begin{equation}
\mathbf{v}(\mathbf{x}_t, t) = \boldsymbol{\epsilon} - \mathbf{x}_0.
\end{equation}
The model predicts $\hat{\mathbf{v}}_\theta(\mathbf{x}_t, t \mid \mathcal{C})$ and is optimized via
\begin{equation}
\mathcal{L}_{\mathrm{FM}} =
\mathbb{E}_{\mathbf{x}_0, \boldsymbol{\epsilon}, t}
\left[
\left\|
\hat{\mathbf{v}}_\theta(\mathbf{x}_t, t \mid \mathcal{C})
-
\mathbf{v}(\mathbf{x}_t, t)
\right\|_2^2
\right].
\end{equation}
During training, the VAE encoder/decoder and ControlNet encoder are frozen, and only lightweight LoRA layers inserted into the DiT backbone are fine-tuned.

\subsection{Training-Free Temporal Consistency Module}
\label{sec:temporal_module}

Although the image generation module produces temporally aligned frames in most cases, minor high-frequency artifacts may still appear, such as jitter around clothing wrinkles or motion boundaries, as illustrated in \cref{fig:method_temporal}. To address these issues without additional training, we introduce a training-free temporal consistency module that refines generated frames directly in the latent diffusion process at inference time. The module consists of two components: low-noise re-denoising and dynamic spatiotemporal regularization.

\paragraph{Low-Noise Re-Denoising.}
Given the latent representations encoded by the VAE, we inject a small amount of Gaussian noise and restart diffusion from a low-noise timestep, following the spirit of SDEdit. Since the underlying diffusion backbone already encodes strong spatiotemporal priors, re-denoising from a lightly perturbed state allows the model to correct minor temporal artifacts while preserving identity, appearance, and motion. Unlike full re-synthesis from pure noise, this refinement primarily targets high-frequency inconsistencies without altering global structure.

\paragraph{Dynamic Spatiotemporal Regularization.}
During re-denoising, we further stabilize temporal dynamics via spatiotemporal low-pass filtering in latent space. The regularization is applied only during the first $35\%$ of inference steps, when global motion patterns are established.

Let $\mathbf{h}_s \in \mathbb{R}^{C \times T \times H \times W}$ denote the latent at step $s$. We apply a three-dimensional Fourier transform over $(T, H, W)$ and attenuate high-frequency components with a smooth Gaussian filter:
\begin{equation}
\tilde{\mathcal{F}}(\mathbf{h}_s) =
\mathcal{F}(\mathbf{h}_s) \odot
\exp\!\left(
- \left(\frac{f_t}{\tau_t}\right)^2
- \left(\frac{f_x}{\tau_s}\right)^2
- \left(\frac{f_y}{\tau_s}\right)^2
\right),
\end{equation}
where $(f_t, f_x, f_y)$ are temporal and spatial frequencies. We set $\tau_t = 0.06$ and $\tau_s = 0.12$ to apply stronger temporal smoothing while preserving spatial details.

The filtered latent is recovered via
\begin{equation}
\tilde{\mathbf{h}}_s = \mathcal{F}^{-1}\!\left(\tilde{\mathcal{F}}(\mathbf{h}_s)\right),
\end{equation}
and used for subsequent denoising. To avoid identity drift, the first-frame latent is treated as a temporal anchor and kept unchanged during filtering.

\section{Experiments}
\subsection{Implementation Details}
We fine-tune FLUX.1 Kontext~\cite{labs2025flux} using LoRA with rank 128. Training is performed on 4 NVIDIA A100 GPUs for 10 epochs, with a batch size of 32. We use the Adam optimizer with a learning rate of $1\times10^{-4}$.

\begin{figure}[t]
    \centering
    \includegraphics[width=0.98\linewidth]{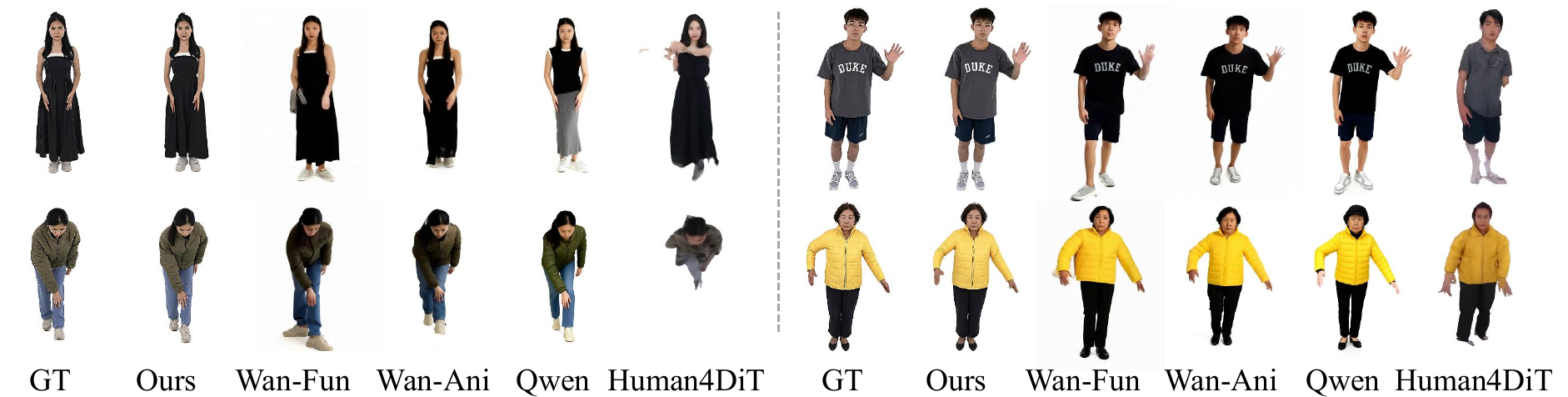}
    \caption{Qualitative comparison for image-to-video human synthesis on the MVHumanNet++ dataset~\cite{li2025mvhumannet++}. We compare our method with Wan-Fun~\cite{Wan2.1FunV1.1_14B_Control}, Wan-Animate (Wan-Ani)~\cite{cheng2025wananimateunifiedcharacteranimation}, Qwen~\cite{wu2025qwen}, and Human4DiT~\cite{shao2024human4dit}. The ground truth (GT) is shown in the first column. }
    \label{fig:comparison_mvhuman}
\end{figure}
\begin{figure}[ht]
    \centering
    \includegraphics[width=0.95\linewidth]{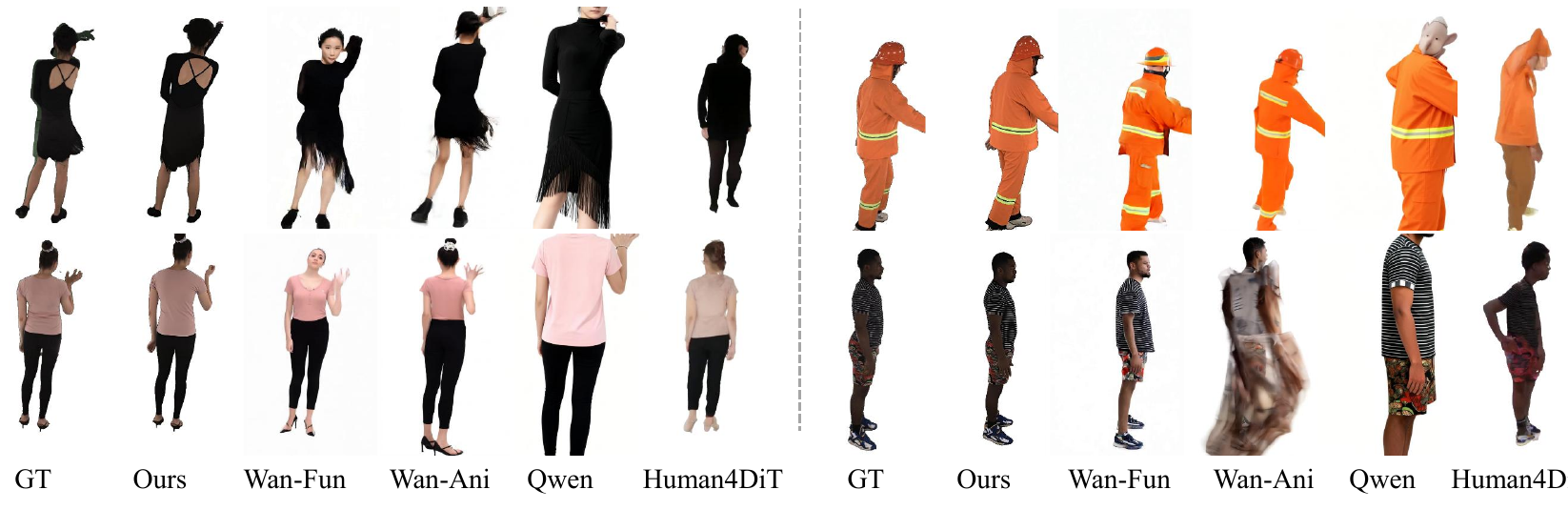}
    \caption{Qualitative comparison on the DNA-Rendering dataset~\cite{Cheng_2023_ICCV}. Our method is evaluated in a zero-shot setting without training on this dataset, demonstrating strong generalization under more challenging viewpoints}
    \label{fig:comparison_dna}
    \vspace{-15px}
\end{figure}

For pose-guided conditioning, we adopt the Flux.1-dev-Controlnet-Surface-Normals checkpoint~\cite{flux1_controlnet_normals}. The ControlNet is kept frozen during training, and its parameters are not updated. For video generation, we employ the Wan 2.1 backbone~\cite{wan2025wan}, specifically the I2V-14B-480P variant. This model is used in a training-free manner, and no parameters are optimized. The number of diffusion inference steps is set to 20 for both the image synthesis module and the training-free temporal consistency module. During inference of the temporal module, we set the denoising strength to 0.7, where values closer to 1.0 indicate less injected noise.

\vspace{-10px}
\subsection{Dataset}
Our framework is trained on MVHumanNet++ \cite{li2025mvhumannet++}, a subset of MVHumanNet \cite{Xiong_2024_CVPR} with higher annotation accuracy. We use 5,000 subjects with 4 views (near front, back, left, and right) to train our proposed method. 
We further evaluate our method on the DNA-Rendering dataset~\cite{Cheng_2023_ICCV}. 
Since the model is not trained on this dataset, the evaluation is conducted in a zero-shot setting using 15 subjects.

\vspace{-10px}
\subsection{Baselines and Evaluation Metrics}
To demonstrate the effectiveness of our proposed method, we compare it against four state-of-the-art baselines, covering both instruction-based image editing and conditional video generation paradigms: Qwen-Image-Edit~\cite{wu2025qwen}, Wan-Animate~\cite{cheng2025wan}, Wan-Fun-Control~\cite{Wan2.1FunV1.1_14B_Control}, and Human4DiT~\cite{shao2024human4dit}.

\textbf{Qwen-Image-Edit} is a powerful instruction-based image editing model capable of performing precise in-painting and generation following textual prompts. \textbf{Wan-Animate} and \textbf{Wan-Fun-Control} represent the state-of-the-art in conditional video generation. Specifically, \textbf{Wan-Animate} specializes in identity-preserving character animation, whereas \textbf{Wan-Fun-Control} excels at structural guidance and motion control via diverse signal inputs. \textbf{Human4DiT} is a pioneering framework for 4D human generation. 

To ensure a consistent evaluation protocol, all methods are provided with the same type of conditioning signals. 
Specifically, the ground-truth first frame is used as the appearance reference, and motion is specified by pose sequences. 
For Human4DiT, motion is given as view-dependent SMPL-X sequences along with a DWPose \cite{yang2023effective} sequence, while for the Wan-family models and Qwen-Image-Edit we construct control videos composed of OpenPose keypoints projected from target camera views.

To assess synthesis quality, we employ five standard metrics covering spatial fidelity, perceptual realism, and temporal coherence. 
PSNR and SSIM measure pixel accuracy and structural similarity between generated frames and ground truth. 
LPIPS evaluates perceptual similarity in deep feature space, while FID measures distribution-level image quality. 
FVD assesses temporal coherence and video dynamics, where lower values indicate motions closer to real videos.

\begin{table*}[t]
\centering
\caption{Quantitative comparison on \textbf{DNA-Rendering} and \textbf{MVHumanNet}. 
\textbf{Bold} indicates the best performance. 
$\uparrow$ means higher is better, $\downarrow$ means lower is better.}
\vspace{-3mm}
\label{tab:combined_comparison}

\setlength{\tabcolsep}{4pt}

\resizebox{\linewidth}{!}{
\begin{tabular}{lccccc@{\hspace{10pt}}ccccc}
\toprule
& \multicolumn{5}{c}{\textbf{DNA-Rendering}} 
& \multicolumn{5}{c}{\textbf{MVHumanNet}} \\
\cmidrule(lr){2-6} \cmidrule(lr){7-11}
\textbf{Method} 
& PSNR$\uparrow$ & SSIM$\uparrow$ & LPIPS$\downarrow$ & FID$\downarrow$ & FVD$\downarrow$
& PSNR$\uparrow$ & SSIM$\uparrow$ & LPIPS$\downarrow$ & FID$\downarrow$ & FVD$\downarrow$ \\
\midrule
\textbf{ReImagine (Ours)} 
& \textbf{22.98} & \textbf{0.847} & \textbf{0.191} & \textbf{57.79} & \textbf{0.561}
& \textbf{23.99} & 0.827 & \textbf{0.165} & \textbf{36.23} & \textbf{0.275} \\

Qwen 
& 18.51 & 0.744 & 0.381 & 100.30 & 1.517
& 19.51 & \textbf{0.831} & 0.182 & 46.33 & 1.442 \\

Wan-Ani 
& 20.55 & 0.802 & 0.267 & 78.92 & 0.689
& 20.72 & 0.829 & 0.193 & 58.89 & 0.403 \\

Wan-Fun 
& 19.29 & 0.822 & 0.244 & 71.09 & 0.637
& 20.83 & 0.827 & 0.215 & 64.69 & 0.401 \\

Human4DiT 
& 16.83 & 0.814 & 0.275 & 74.96 & 0.833
& 19.37 & 0.822 & 0.258 & 70.64 & 0.574 \\

\bottomrule
\end{tabular}
}
\vspace{-15px}
\end{table*}

\vspace{-10px}
\subsection{Quantitative and Qualitative Analysis}
\vspace{-5px}
\cref{fig:comparison_mvhuman} compares our method with several state-of-the-art baselines on MVHumanNet++. 
Under frontal views, most methods generate visually plausible results; however, our method produces slightly sharper details and more consistent appearances than Wan-Fun, Wan-Animate, and Qwen.
When pose changes become larger, our method remains the most stable, while Wan-Animate and Qwen remain partially stable but already exhibit artifacts or structural inconsistencies.
\cref{fig:comparison_dna} further evaluates performance under more challenging viewpoint changes. 
In this setting, our method produces more consistent human appearances and structures, whereas baselines often suffer from identity drift or degraded geometry. 
Among them, Wan-Fun and Wan-Animate occasionally produce reasonable results, Qwen degrades noticeably, and Human4DiT fails in most cases. More results are provided in the supplementary materials.

\begin{figure}[t]
\centering

\begin{minipage}[t]{0.4\linewidth}
  \centering
   \vspace{-160pt}
  \captionsetup{type=table, justification=raggedright, singlelinecheck=false, width=\linewidth}
  \captionof{table}{Aesthetic quality results evaluated using VBench~\cite{huang2023vbench}. We compare four strategies: \textbf{RD+3DFFT (ours)}, low-noise \textbf{R}e-\textbf{D}enoising with 3DFFT-based spatiotemporal regularization; \textbf{RD}, re-denoising only; \textbf{RD+Med}, re-denoising with temporal median filtering; and \textbf{IF}, \textbf{I}mage-\textbf{F}irst generation without temporal refinement.}
  \label{fig:ablation_vis}

  \vspace{2pt}
  \small
  \setlength{\tabcolsep}{4pt}
  \renewcommand{\arraystretch}{1.05}
  \begin{tabular}{l|c}
    \hline
    Strategy & Aesthetic Score $\uparrow$ \\ \hline
    \textbf{RD+3DFFT}  & \textbf{0.5346} \\
    RD & 0.5326 \\
    RD+Med  & 0.5338 \\
    IF  & 0.5303 \\ \hline
  \end{tabular}
\end{minipage}
\hfill
\begin{minipage}[t]{0.58\linewidth}
  \centering
  \includegraphics[width=\linewidth]{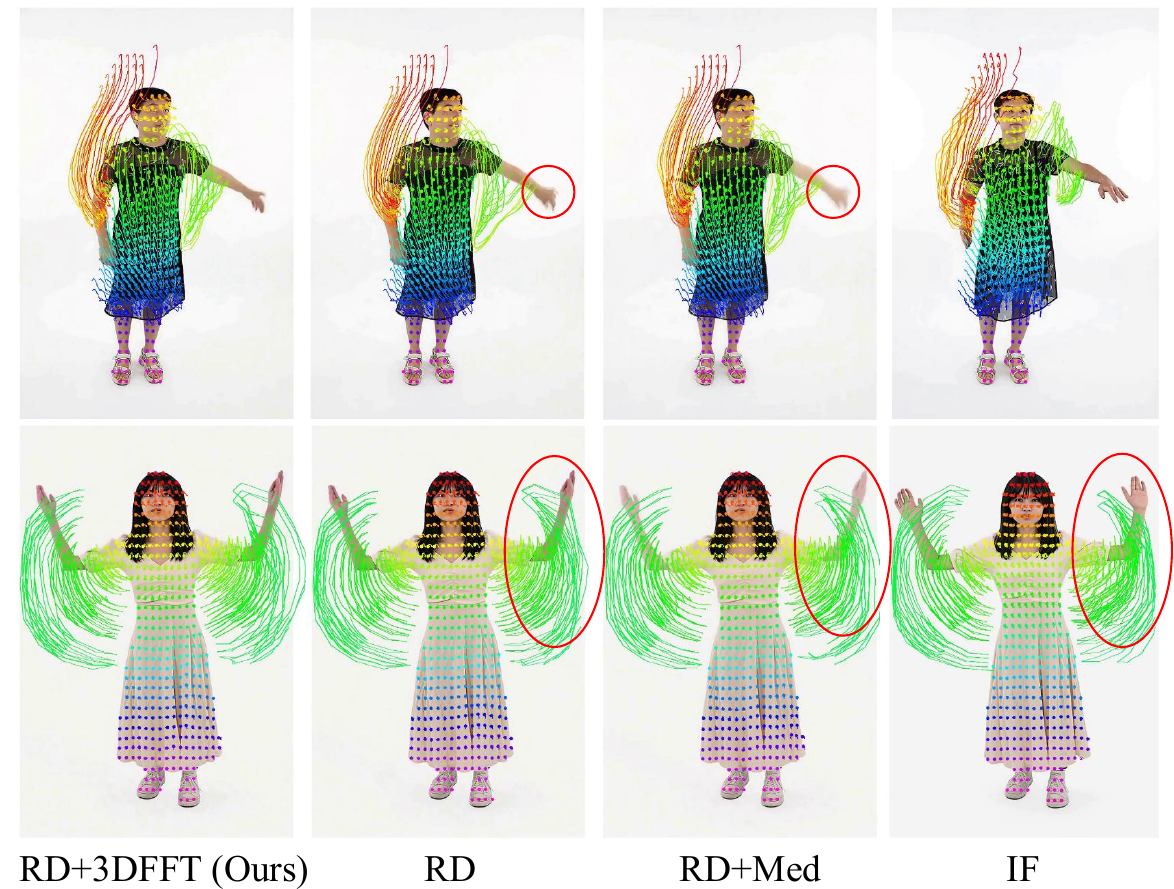}
  \captionsetup{type=figure, justification=raggedright, singlelinecheck=false, width=0.98\linewidth}
  \captionof{figure}{Temporal consistency ablation via tracking visualization.}
  \label{tab:ablation}
\end{minipage}
\vspace{-10px}
\end{figure}

\cref{tab:combined_comparison} reports quantitative results on DNA-Rendering and MVHumanNet. 
While recent video generation models perform well in general settings, they struggle in the task of generating view-consistent 360-degree human videos. 
The large gaps in FID and FVD indicate that baseline methods fail to maintain consistent appearances under unseen viewpoints (e.g., side and back views), often producing artifacts and structural distortions as the camera rotates.

In contrast, our method achieves substantially better temporal consistency, obtaining an FVD of 0.275 on MVHumanNet compared to 0.403 for Wan-Animate. 
Although Qwen-Image-Edit attains a relatively high SSIM (0.831), this metric is misleading for video generation: it produces high-quality individual frames but lacks temporal coherence, resulting in a much worse FVD of 1.442. 
Human4DiT also struggles to generalize to the complex clothing and appearance variations in our evaluation set.

\vspace{-10px}
\subsection{Ablation Study}

\paragraph{Temporal Consistency Ablation via Tracking Visualization.}
We compare temporal consistency strategies using tracking visualization with CoTracker3~\cite{karaev24cotracker3} (\cref{fig:ablation_vis}), including Image-First generation without refinement, low-noise re-denoising only, re-denoising with spatiotemporal spectral regularization, and re-denoising with median filtering.

Without temporal refinement, trajectories are fragmented and unstable, particularly in fast-moving regions such as arms (red). Re-denoising alleviates some artifacts but retains jitter, while median filtering oversmooths and drifts. In contrast, our 3DFFT-based regularization produces the smoothest and most coherent trajectories. Although removing the Wan backbone~\cite{wan2025wan} yields sharp individual frames, it results in poor temporal continuity and incorrect motion, underscoring the importance of strong video priors and structured temporal regularization. Video results are provided in the supplementary materials.

\begin{figure}[t]
\centering
\begin{minipage}[t]{0.63\linewidth}
  \centering
  \includegraphics[width=0.98\linewidth]{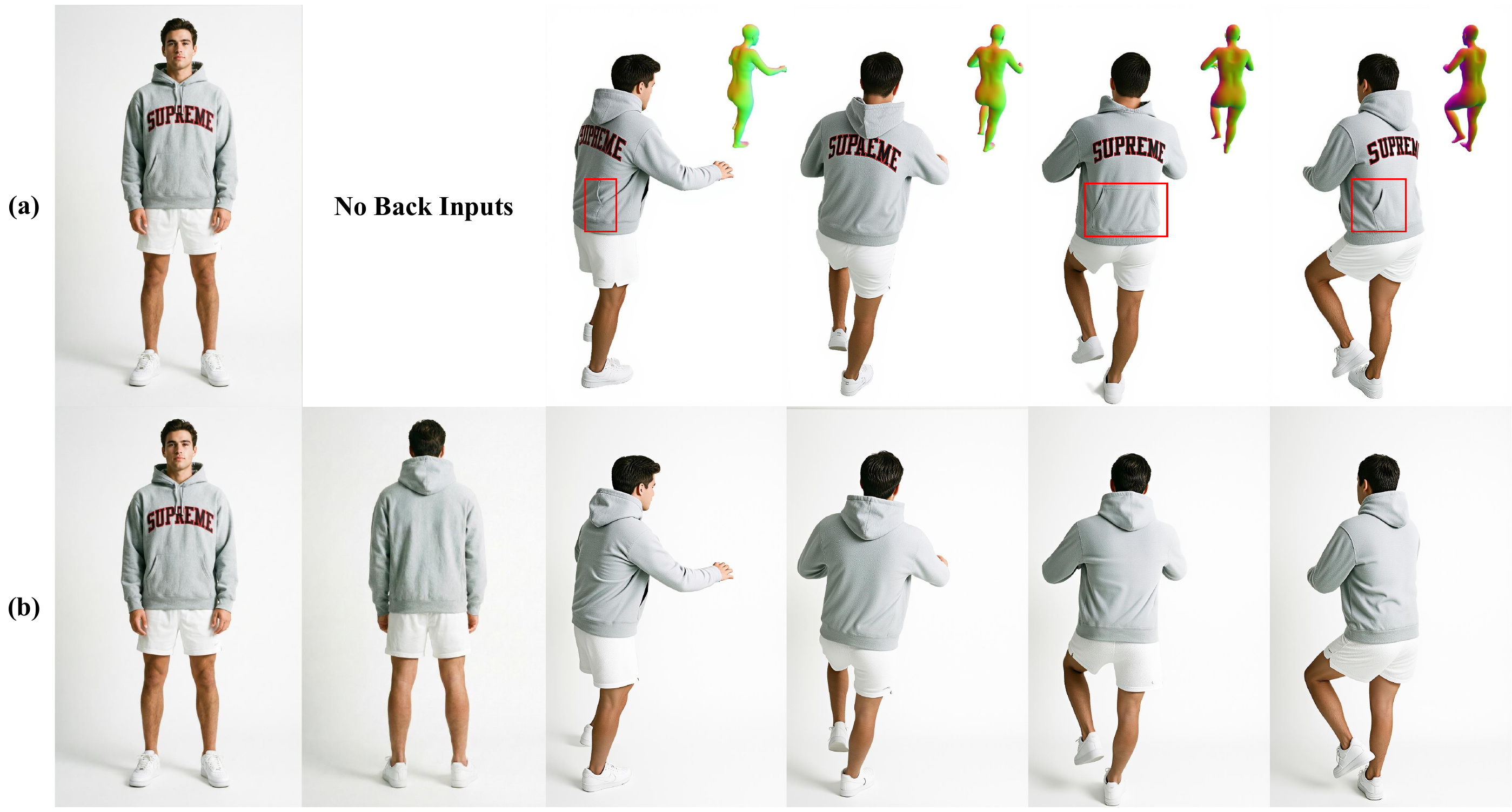}
  \captionsetup{type=figure, justification=raggedright, singlelinecheck=false, width=0.98\linewidth}
  \captionof{figure}{Ablation on missing back-view appearance input. 
The two leftmost columns show the input appearance images, where the back-view input is replaced by a blank image in row (a).}
  \label{fig:ablation_back_view}
\end{minipage}
\hfill
\begin{minipage}[t]{0.34\linewidth}
    \centering
    \vspace{-105pt}
    \captionsetup{type=table, justification=raggedright, singlelinecheck=false}
    \captionof{table}{User study (\% pairwise preference). 
\textbf{View.}: view consistency; 
\textbf{Temp.}: temporal smoothness. Higher is better.}
    \label{tab:user_study}
    \vspace{2pt}
    
    \footnotesize
    \setlength{\tabcolsep}{3pt}
    \renewcommand{\arraystretch}{1.0}
    \resizebox{\linewidth}{!}{
    \begin{tabular}{lcc}
    \toprule
    Method & View. & Temp. \\
    \midrule
    Qwen        & 15.6 & 16.2 \\
    Wan-Ani     & 24.1 & 24.5 \\
    Wan-Fun     & 26.8 & 26.8 \\
    Human4DiT   & 11.7 & 21.3 \\
    \textbf{ReImagine} & \textbf{41.8} & \textbf{34.7} \\
    \bottomrule
    \end{tabular}
    }

\end{minipage}

\vspace{-5px}
\end{figure}

\paragraph{Ablation on Missing Back-View Appearance.}
We study an extreme setting where the back-view appearance input is replaced by a blank image (\cref{fig:ablation_back_view}), evaluating the robustness of our image-first framework without explicit back-view information. The leftmost columns show the input appearances, followed by synthesized results under different target viewpoints.

Even without back-view input, the model generates visually plausible results, indicating partial inference of unseen appearance from front-view input and motion guidance. However, controllability degrades for back-facing views, where front-view elements are often propagated to the back (\cref{fig:ablation_back_view} (a)), highlighting the importance of explicit front–back appearance conditioning for accurate view-dependent generation.

\begin{table}[t]
\centering
\caption{Comparison with a video-first baseline that directly trains a video diffusion model using the same canonical appearance inputs and pose sequences. Our image-first formulation achieves significantly better spatial fidelity and temporal stability.}
\resizebox{0.75\linewidth}{!}{
\begin{tabular}{lccccc}
\toprule
\textbf{Method} & PSNR$\uparrow$ & SSIM$\uparrow$ & LPIPS$\downarrow$ & FID$\downarrow$ & FVD$\downarrow$ \\
\midrule
Video-first (Uni-Animate DiT) & 19.05 & 0.814 & 0.219 & 55.61 & 0.614 \\
\textbf{Image-first (Ours ReImagine)} & \textbf{23.99} & \textbf{0.827} & \textbf{0.165} & \textbf{36.23} & \textbf{0.275} \\
\bottomrule
\end{tabular}
}
\label{tab:video_first_baseline}
\vspace{-15px}
\end{table}

\vspace{-10px}

\subsection{Discussion with Video-First Training}
Our method adopts an \emph{image-first} formulation for controllable human video generation. 
An alternative design is to directly train a video generation model conditioned on the same inputs, including canonical front--back appearance references and SMPL-X motion sequences. 
To better understand the advantages of the proposed formulation, we implement such a video-first baseline and compare it with our approach.

\paragraph{Video-first baseline.}
We construct a baseline based on the Uni-Animate DiT \footnote{\url{https://github.com/ali-vilab/UniAnimate-DiT}} architecture with the Wan video diffusion backbone. 
The model receives the same inputs as our method, namely canonical front--back reference images and SMPL-X pose, and is trained on the same MVHumanNet++ dataset. 
This setup ensures a fair comparison: both methods use the same inputs and training data, and only differ in the generation strategy (video-first vs.\ image-first).

\begin{figure}[t]
    \vspace{-10px}
    \centering
    \includegraphics[width=0.92\linewidth]{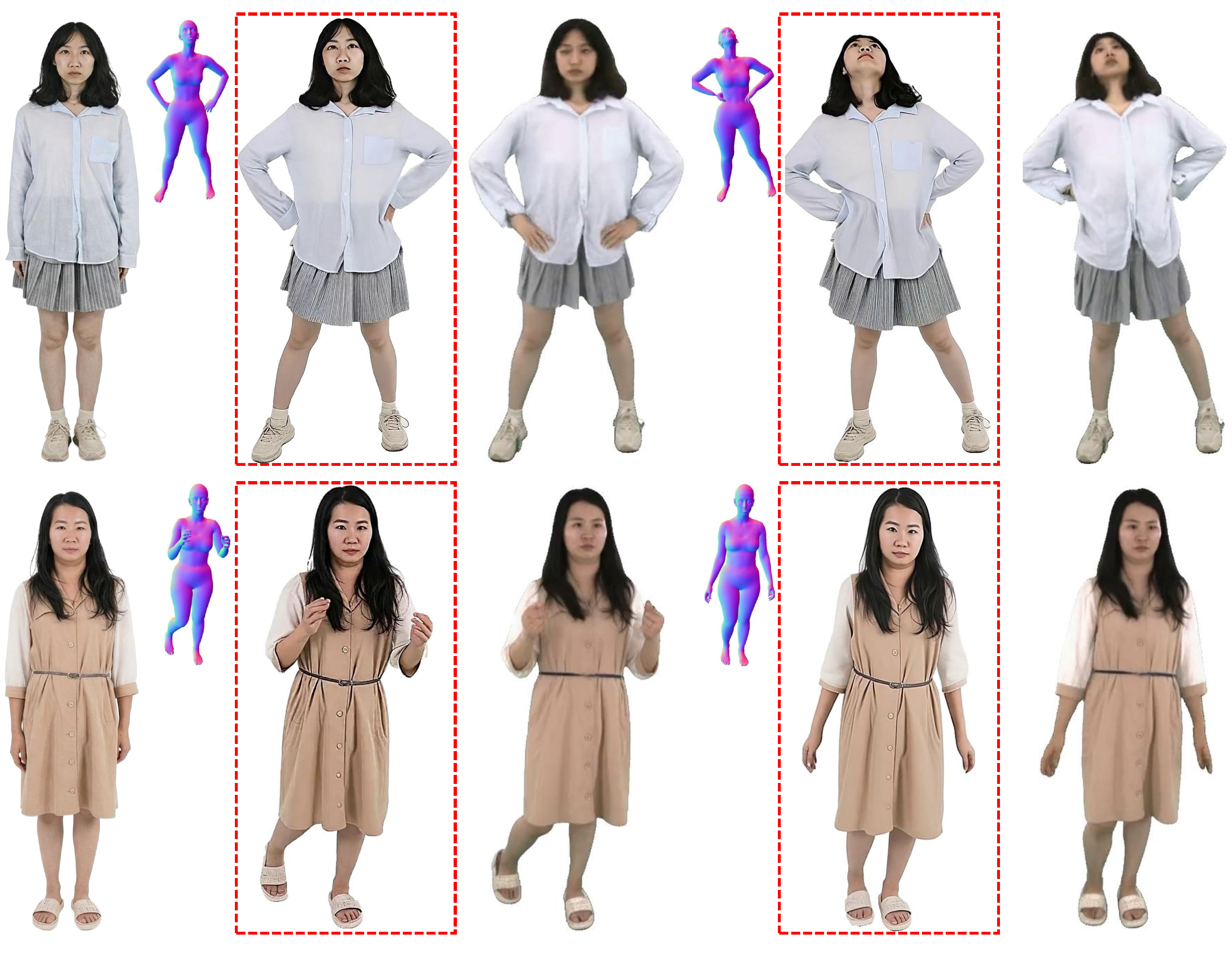}
\caption{
Qualitative comparison between our image-first method and a video-first baseline (Uni-Animate DiT).
The leftmost column shows the canonical front reference input (back reference omitted for space).
Each row corresponds to the same subject under different poses.
Red dashed boxes highlight results generated by our image-first pipeline, while the remaining images are produced by the video-first baseline.
Our method preserves sharper textures and clearer facial and hands details.
}
    \label{fig:video_first_baseline}
    \vspace{-20px}
\end{figure}

\paragraph{Empirical comparison.}
As shown in \cref{tab:video_first_baseline}, the video-first baseline performs consistently worse across all quantitative metrics.
In particular, our method achieves higher PSNR and SSIM while significantly reducing LPIPS, FID, and FVD, indicating improved spatial fidelity and temporal stability.
Qualitative comparisons in \cref{fig:video_first_baseline} further reveal that our method produces noticeably sharper textures, more faithful color reproduction, and clearer facial details.
In contrast, the video-first baseline tends to generate blurrier appearances and degraded fine structures.

\paragraph{Analysis.}
We attribute this gap primarily to the limited availability and quality of multi-view human video data. 
Training a generalizable video diffusion model typically requires large-scale, high-quality datasets with consistent multi-view supervision, which remain scarce in the human generation domain. 
Consequently, video models trained on such data often inherit the visual characteristics of the training distribution, limiting the achievable perceptual fidelity.

By contrast, our image-first formulation leverages the strong generative priors of modern image diffusion backbones trained on large-scale and high-quality image dataset.
This enables higher-quality spatial synthesis while reducing the reliance on large multi-view video datasets.
Under this formulation, the temporal stage only needs to enforce motion coherence across independently generated frames, rather than learning full video synthesis from limited video data.

\paragraph{Discussion.}
These results support our motivation to formulate controllable human video generation as a motion-guided image synthesis problem.
By relying on strong image generation priors and lightweight temporal stabilization, the proposed approach achieves higher visual fidelity while maintaining temporal coherence.

\begin{figure}[t]    
    \centering
    \includegraphics[width=0.98\linewidth]{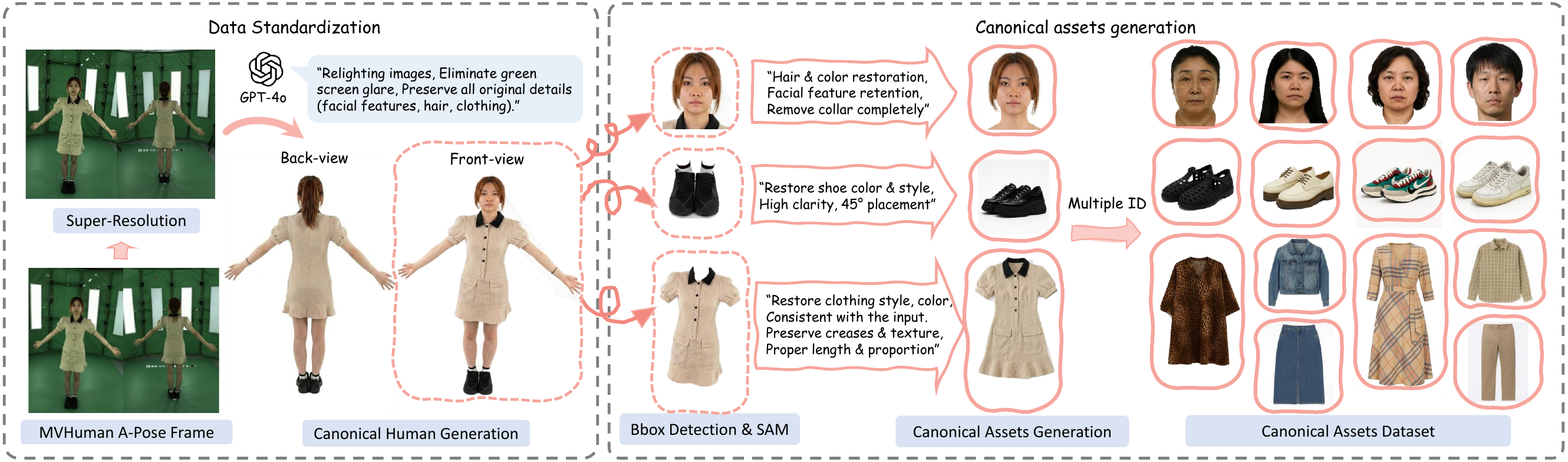}
    \caption{Data construction pipeline for building canonical and disentangled human assets from MVHumanNet \cite{Xiong_2024_CVPR}}
    \label{fig:data_pipeline}
\end{figure}
\begin{figure}[t]
    \centering
    \includegraphics[width=0.98\linewidth]{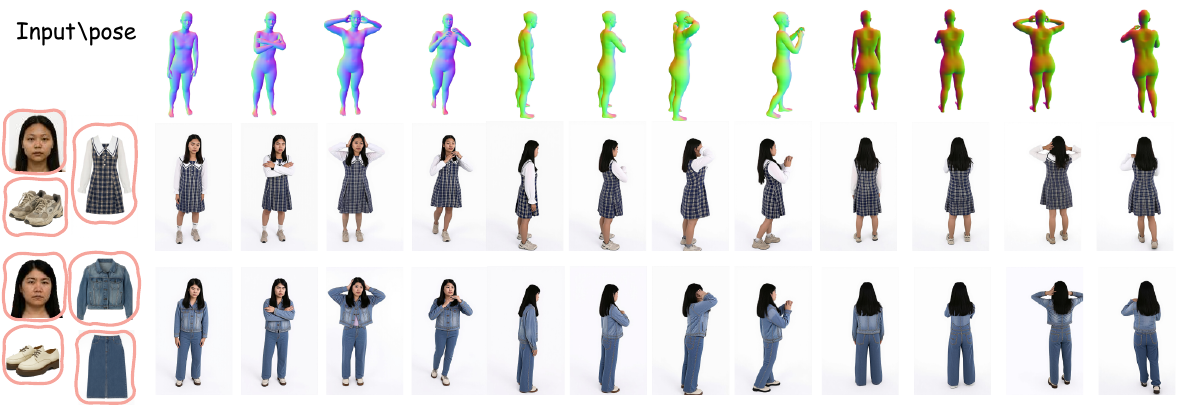}
    \caption{More results produced by our end-to-end method.}
    \label{fig:supp_1}
    \vspace{-15px}
\end{figure}

\section{User Study}
\vspace{-5px}
We conduct a user study to compare our method with Qwen, Wan-Animate, Wan-Fun, and Human4DiT. 
For each method, human videos are generated using the same appearance reference (the ground-truth first frame) and comparable pose sequences. 
Participants perform pairwise comparisons on randomly ordered results, and the preference rates are aggregated over all comparisons.

\vspace{-10px}
\section{End-to-End Extension with Canonical Asset Data}
\vspace{-5px}
As a complementary extension, we further relax the input requirements of our framework by constructing a canonical asset dataset that enables end-to-end human image synthesis from disentangled components such as face, clothing, and shoes.

\paragraph{Data Construction Pipeline.}
The pipeline is illustrated in \cref{fig:data_pipeline}. Starting from MVHumanNet~\cite{Xiong_2024_CVPR}, we extract front- and back-view frames in canonical A-poses and enhance them using the super-resolution model HYPIR~\cite{lin2025harnessing}. GPT-4o~\cite{hurst2024gpt} is then used to remove environmental lighting and relight the subject under neutral illumination, producing standardized human images.

Next, YOLO-World~\cite{cheng2024yolo} detects component regions and SAM~\cite{Kirillov_2023_ICCV} performs segmentation to extract assets such as faces, garments, and shoes. These components are further canonicalized using GPT-4o to obtain front-facing faces, clean garments with preserved textures, and standardized shoes. After large-scale augmentation and manual filtering, the final dataset contains approximately 1,600 identities with disentangled canonical assets.

\paragraph{End-to-End Training with Disentangled Assets.}
Using the constructed dataset, we train an end-to-end model by reusing the image generation module described in \cref{sec:image_module}. Specifically, we replace the original front-view and back-view human image inputs with disentangled face, clothing, and shoe assets, each injected into the model with appropriate condition-aware positional encoding. The rest of the architecture and training procedure remain unchanged. This extension demonstrates that our framework naturally generalizes to more flexible input representations and supports end-to-end synthesis from canonical, user-friendly assets, as illustrated in \cref{fig:supp_1}. More examples are provided in the supplementary materials.

\section{Conclusion}
We present an image-first approach to human video generation, based on the insight that generating high-quality human images under pose and viewpoint conditioning is the primary challenge. By leveraging pretrained image synthesis backbones and structured pose guidance from SMPL-X, our method synthesizes temporally coherent human videos without requiring high-quality video data or video-specific training. We additionally release a canonical dataset and an end-to-end variant enabled by synthetic canonical data, providing complementary resources for future research on human generation.


%
%
\bibliographystyle{splncs04}
\bibliography{main}

\clearpage
\begin{nolinenumbers}
\begin{center}
{\Large \textbf{Supplementary Material}}
\end{center}
\end{nolinenumbers}

\appendix

This document provides additional discussion, experimental details, ablation studies, and qualitative results that complement the main paper.

\section{Extended Temporal Consistency Evaluation}

We further analyze the temporal consistency module through additional quantitative comparisons.
Several temporal strategies are evaluated on top of the image-first pipeline, reporting both spatial quality metrics and temporal stability metrics.

\vspace{-5px}
\begin{table}[ht]
\centering
\caption{
Comparison of different temporal strategies in terms of spatial image quality.
}
\vspace{-5px}
\resizebox{0.75\linewidth}{!}{
\begin{tabular}{lccccc}
\toprule
\textbf{Method} & PSNR$\uparrow$ & SSIM$\uparrow$ & LPIPS$\downarrow$ & FID$\downarrow$ & FVD$\downarrow$ \\
\midrule
Image-first w/o temporal module & \textbf{24.21} & \textbf{0.831} & \textbf{0.158} & \textbf{34.87} & \textbf{0.261} \\
Re-Denoising & 24.05 & 0.829 & 0.162 & 35.94 & 0.268 \\
Re-Denoising + Median Filter & 23.92 & 0.828 & 0.164 & 36.78 & 0.272 \\
Re-Denoising + 3D FFT (ours) & 23.99 & 0.827 & 0.165 & 36.23 & 0.275 \\
\bottomrule
\end{tabular}
}
\label{tab:temp_ablation_methods}
\vspace{-25px}
\end{table}

\begin{table}[ht]
\centering

\caption{
Temporal consistency evaluation using a flow-based metric.
}
\vspace{-5px}
\resizebox{0.55\linewidth}{!}{
\begin{tabular}{lc}
\toprule
\textbf{Method} & Flow Error$\downarrow$ \\
\midrule
Image-first w/o temporal module  & 0.552 \\
Re-Denoising  & 0.616 \\
Re-Denoising + Median Filter  & 0.619 \\
Re-Denoising + 3D FFT (ours)  & \textbf{0.481} \\
\bottomrule
\end{tabular}
}
\label{tab:temp_ablation_temporal}
\vspace{-10px}
\end{table}

\cref{tab:temp_ablation_methods} reports spatial image quality metrics.
The image-first pipeline without temporal processing achieves the best per-frame fidelity since each frame is generated independently.
Introducing temporal refinement slightly decreases spatial metrics, which is expected because temporal smoothing suppresses high-frequency details.
Among the tested strategies, the proposed 3D FFT approach maintains competitive image quality while improving temporal coherence.

\cref{tab:temp_ablation_temporal} evaluates temporal stability using a flow-based metric.
The proposed 3D FFT module significantly reduces the optical-flow reconstruction error, indicating improved frame-to-frame consistency.

For the flow-based metric, we compute an optical-flow-guided\footnote{We use this implementation: \url{https://github.com/facebookresearch/co-tracker}} warping error between consecutive frames.
Given frames $I_t$ and $I_{t+1}$, we estimate the optical flow $F_{t \rightarrow t+1}$ and warp $I_t$ toward the next frame:
\vspace{-5px}
\[
\hat{I}_{t+1} = \mathcal{W}(I_t, F_{t \rightarrow t+1}),
\]
where $\mathcal{W}(\cdot)$ denotes differentiable image warping.
The temporal error is computed as
\vspace{-5px}
\[
E_{\mathrm{flow}} =
\frac{1}{T-1}\sum_{t=1}^{T-1}
\left\|
\hat{I}_{t+1} - I_{t+1}
\right\|_1.
\]
Lower values indicate better temporal alignment between adjacent frames.
Additional qualitative comparisons are provided in the supplementary video.

\section{Additional Qualitative Results}
\vspace{-10px}
\begin{figure}[ht]
    \centering
    \includegraphics[width=0.98\linewidth]{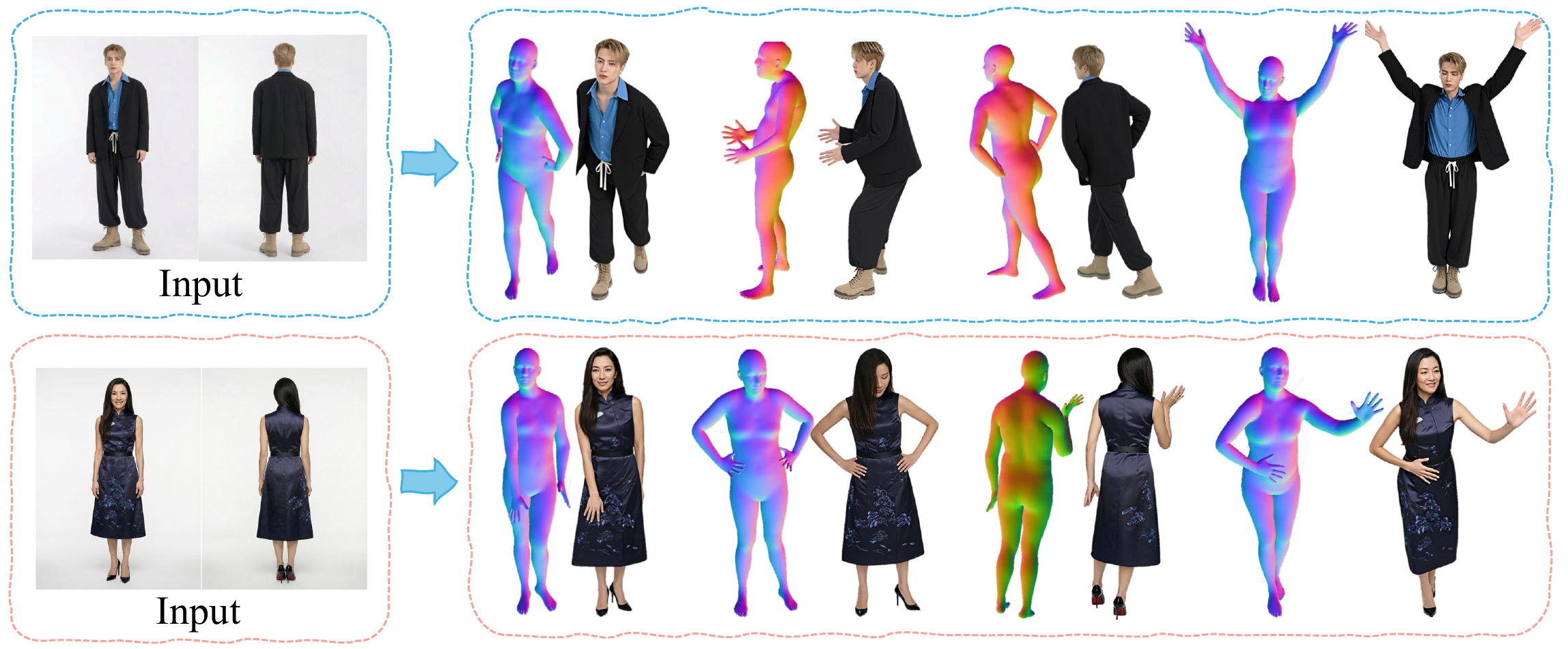}
    \caption{
Qualitative results on in-the-wild appearance inputs.
Given a single reference image and SMPL-X pose sequences, our model generates consistent human images under diverse poses and viewpoints while preserving identity and clothing details.
}
    \label{fig:supp_wild}
    \vspace{-20px}
\end{figure}
\begin{figure}[h]
    \centering
    \includegraphics[width=0.98\linewidth]{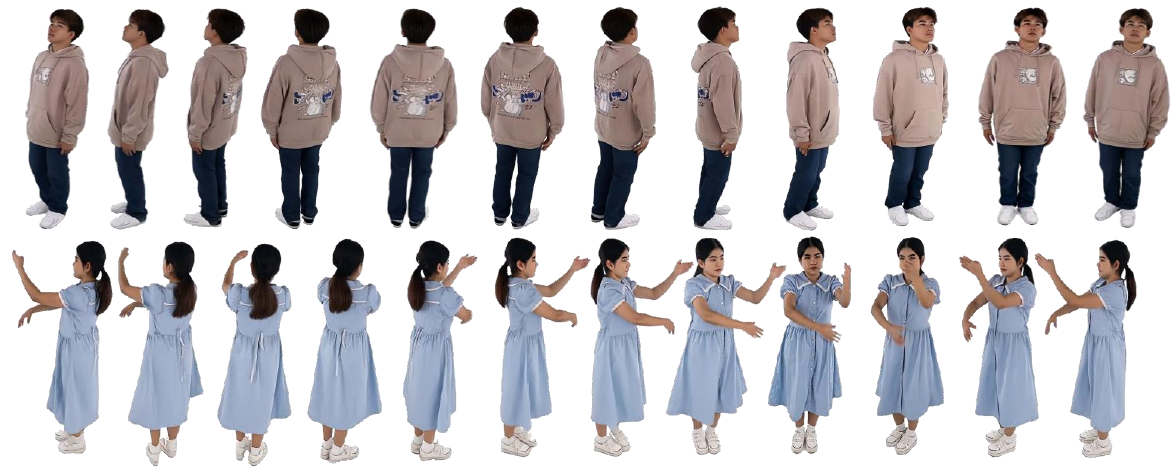}
    \caption{
Qualitative results under large viewpoint changes.
Starting from a canonical appearance input, the model synthesizes consistent human images as the camera viewpoint rotates around the subject, demonstrating strong control over viewpoint and pose.
}
    \label{fig:supp_rotation}
    \vspace{-20px}
\end{figure}

We present additional qualitative results to further demonstrate the generalization ability and controllability of our framework. 
As shown in \cref{fig:supp_wild}, our model can robustly handle in-the-wild appearance inputs and synthesize consistent human images under diverse poses and viewpoints. 
Despite the variability in real-world appearances, the generated results maintain coherent identity and stable visual details, indicating that the model generalizes well beyond the controlled training scenarios.

We further evaluate the model under large viewpoint changes. 
As illustrated in \cref{fig:supp_rotation}, our method produces stable results under turntable-style viewpoint variations, where the camera rotates around the subject while preserving appearance consistency. 
These examples highlight the effectiveness of the proposed disentangled conditioning mechanism for controlling both pose and viewpoint.

For a more comprehensive evaluation, additional qualitative results and full video sequences are provided in the supplementary video, where the temporal behavior and viewpoint transitions can be better observed.

\section{Additional Applications with Canonical Asset Training}
\begin{figure}[t]
    \vspace{-5px}
    \centering
    \includegraphics[width=0.98\linewidth]{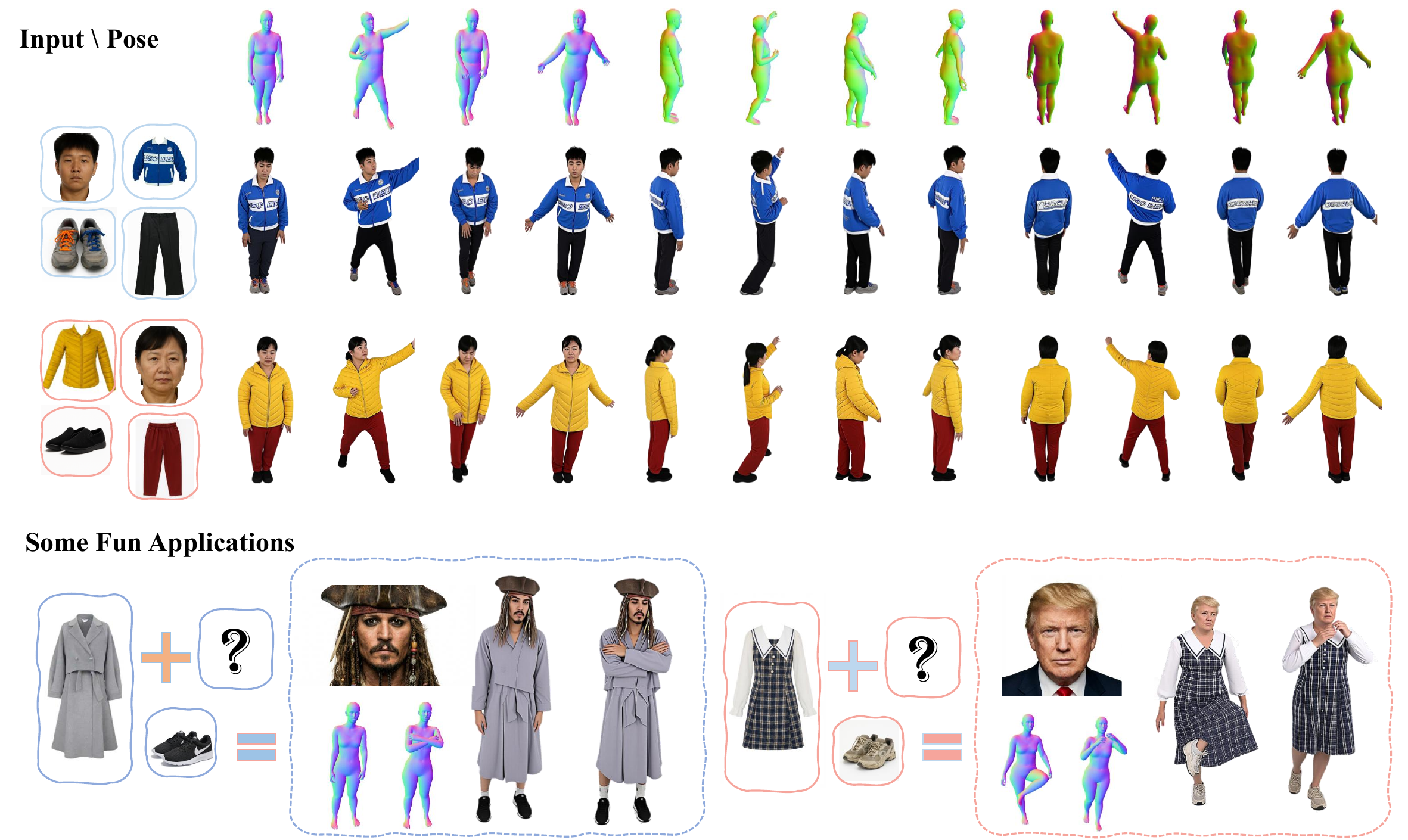}
    \caption{
Examples of additional applications enabled by our canonical asset-based training.
The model can synthesize humans under diverse poses, combine identity and clothing assets for creative generation, and control pose under arbitrary viewpoints.
}
    \label{fig:supp_2}
    \vspace{-15px}
\end{figure}

\begin{table}[t]
\vspace{-5px}
\centering
\caption{
Quantitative reports on our image-first pipeline and a disentangled asset-based pipeline that uses separate face, clothing, and shoe inputs with pose conditioning.
While the disentangled setting provides a more flexible and user-friendly interface, it introduces a more challenging generation task, leading to slightly lower quantitative performance.
}
\resizebox{0.75\linewidth}{!}{
\begin{tabular}{lcccc}
\toprule
\textbf{Method} & PSNR$\uparrow$ & SSIM$\uparrow$ & LPIPS$\downarrow$ & FID$\downarrow$ \\
\midrule
Disentangled Asset Pipeline & 22.74 & 0.821 & 0.178 & 39.81 \\
\textbf{Image-first (Ours ReImagine)} & \textbf{23.99} & \textbf{0.827} & \textbf{0.165} & \textbf{36.23} \\
\bottomrule
\end{tabular}
}
\label{tab:disentangled_pipeline}
\vspace{-10px}
\end{table}

Beyond the standard controllable human video synthesis task, the canonical asset-based training formulation also enables several additional applications.
As illustrated in \cref{fig:supp_2}, the model can flexibly compose human appearances from disentangled assets and generate consistent results under diverse poses and viewpoints.

The canonical asset-based training also enables several additional applications.
Given disentangled inputs such as face identity, clothing items, and shoes, the model can compose these elements into a coherent human appearance and generate pose-conditioned results.
Since the assets are injected with condition-aware positional encoding, the model learns to associate each component with its corresponding spatial and semantic role during training.

This disentangled representation further allows flexible appearance recomposition.
Different identities and clothing assets can be freely combined to create novel character appearances that are not explicitly observed during training, enabling a variety of creative generation scenarios.

Moreover, the model supports pose transfer under arbitrary viewpoints.
Starting from canonical appearance inputs, the system can synthesize the subject performing different motions while maintaining appearance consistency across viewpoints.
These results demonstrate that the canonical asset formulation provides a flexible interface for controllable human synthesis and extends beyond standard video generation.

\vspace{-5px}
\section{User Study Details}
\vspace{-5px}
We provide additional details of the user study protocol used in \cref{tab:user_study}. 
The study evaluates perceptual preference across two criteria: \textbf{view consistency} and \textbf{temporal smoothness}.

\textbf{Participants.}
We recruited 30 participants for the study. All participants had normal or corrected-to-normal vision. 
The participants included both researchers familiar with computer vision/graphics and non-expert users.

\textbf{Stimuli.}
For each method, human videos are generated using the same appearance reference (the ground-truth first frame) and identical or comparable SMPL-X pose sequences. 
This ensures that the compared results differ only due to the generation method.

\textbf{Compared Methods.}
The evaluated methods include Qwen, Wan-Animate, Wan-Fun, Human4DiT, and our method.

\textbf{Evaluation Protocol.}
Participants performed pairwise comparisons between the generated videos. 
For each comparison, two videos produced by different methods were presented side-by-side in randomized order to avoid positional bias.

Participants were asked to choose the preferred video under the following criteria:

\begin{itemize}
\item \textbf{View consistency (View.)}: whether the generated subject maintains consistent appearance under viewpoint changes.
\item \textbf{Temporal smoothness (Temp.)}: whether the motion is smooth and free of temporal artifacts such as flickering or identity drift.
\end{itemize}

Each participant evaluated 20 randomly sampled video pairs for each criterion.

\textbf{Aggregation.}
For each method pair, the preference rate is computed as the percentage of times a method is selected over the others. 
The final results reported in \cref{tab:user_study} are aggregated over all comparisons across participants.

\end{document}